\documentclass[10pt,twocolumn,letterpaper]{article}

\usepackage{cvpr}
\usepackage{times}
\usepackage{epsfig}
\usepackage{graphicx}
\usepackage{amsmath}
\usepackage{amssymb}
\usepackage{amsfonts}
\usepackage{array}
\usepackage{algorithm}
\usepackage{algorithmic}
\usepackage{cite}
\usepackage{booktabs}
\usepackage{float}
\usepackage{multirow}

\usepackage[font=small,labelfont=bf,tableposition=top]{caption}
\usepackage{blindtext}
\usepackage[T1]{fontenc}
\usepackage[latin1]{inputenc}
\usepackage[american]{babel}
\usepackage{cuted}
\usepackage{caption}

\newcommand\norm[1]{\left\lVert#1\right\rVert}

\usepackage[pagebackref=true,breaklinks=true,letterpaper=true,colorlinks,bookmarks=false]{hyperref}

\cvprfinalcopy 

\def\cvprPaperID{8592} 

\ifcvprfinal\fi
\begin{document}

\title{Social-WaGDAT: Interaction-aware Trajectory Prediction via Wasserstein Graph Double-Attention Network}

\author{Jiachen Li\qquad Hengbo Ma\qquad Zhihao Zhang\qquad Masayoshi Tomizuka\\
University of California, Berkeley\\
{\tt\small \{jiachen\_li, hengbo\_ma, zhihaozhang, tomizuka\}@berkeley.edu}}

\maketitle

\begin{abstract}
Effective understanding of the environment and accurate trajectory prediction of surrounding dynamic obstacles are indispensable for intelligent mobile systems (like autonomous vehicles and social robots) to achieve safe and high-quality planning when they navigate in highly interactive and crowded scenarios.
Due to the existence of frequent interactions and uncertainty in the scene evolution, it is desired for the prediction system to enable relational reasoning on different entities and provide a distribution of future trajectories for each agent.
In this paper, we propose a generic generative neural system (called Social-WaGDAT) for multi-agent trajectory prediction, which makes a step forward to explicit interaction modeling by incorporating relational inductive biases with a dynamic graph representation and leverages both trajectory and scene context information.
We also employ an efficient kinematic constraint layer applied to vehicle trajectory prediction which not only ensures physical feasibility but also enhances model performance.
The proposed system is evaluated on three public benchmark datasets for trajectory prediction, where the agents cover pedestrians, cyclists and on-road vehicles. The experimental results demonstrate that our model achieves better performance than various baseline approaches in terms of prediction accuracy.
\end{abstract}
\vspace{-0.5cm}

\section{Introduction}
In order to navigate safely in dense traffic scenarios or crowded areas full of vehicles and pedestrians, it is crucial for autonomous vehicles or mobile robots to forecast future behaviors of surrounding interactive agents accurately and efficiently \cite{lefevre2014survey}. For short-term prediction, it may be acceptable to use pure physics based methods. However, due to the uncertain nature of future situations, the system for long-term prediction is desired to not only allow for interaction modeling between different agents, but also to figure out traversable regions delimited by road layouts as well as right of way compliant to traffic rules.
Figure 1 illustrates several traffic scenarios where interaction happens frequently and the drivable areas are heavily defined by road geometries. For instance, at the entrance of roundabouts or unsignalized intersections, the future behavior of an entering vehicle highly depends on whether 
the conflicting vehicles would yield and leave enough space for it to merge. In addition, for vehicle trajectory prediction, kinematic constraints should be satisfied to make the trajectories feasible and smooth. 
\begin{strip}
\centering\noindent
\includegraphics[width=\textwidth]{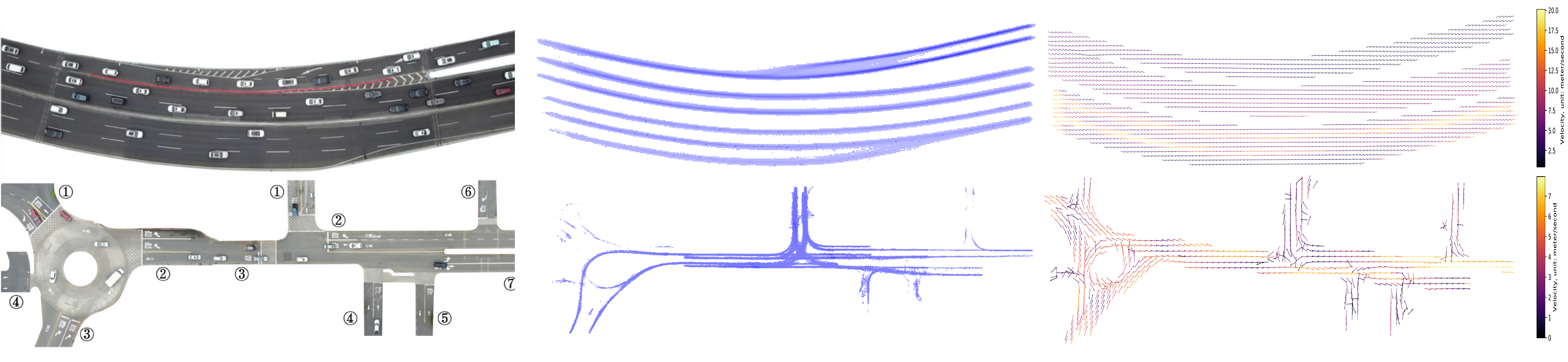}
\captionof{figure}{Typical traffic scenarios with large uncertainty and interactions among multiple entities. The left column is adopted from \cite{interactiondataset}. The upper figure in the first column was captured in a highway ramp merging scenario, where lane change behavior with negotiation happens frequently. The lower figure was captured in a roundabout and an unsignalized intersection scenario, where yielding and stopping behaviors happen frequently. The other two columns shows the occupancy density maps and the velocity fields of the scenarios, which are generated based on the training data to provide statistical context information.}
\end{strip}
There have been extensive studies on the prediction of a single target entity, which consider the influences of its surrounding entities \cite{hong2019rules,chai2019multipath,alahi2016social,gupta2018social}. However, such approaches only care about a one-way interaction, but ignore the potential interactions in the opposite way. Recent works have tried to address this issue by simultaneous forecasting for multiple agents \cite{deo2018convolutional,deo2018would,zhao2019multi}. However, most of these methods just use concatenation or pooling operations to blend the features of different agents without explicit relational reasoning. 
Moreover, they are not able to model higher-order interactions beyond adjacent entities.
In this work, we make a step forward to model the interactions explicitly with a spatio-temporal graph representation and the message passing rules defined by graph networks \cite{battaglia2018relational}, which enables permutation invariance and mutual effects between pairs of entities. Moreover, since it is also essential to figure out which of the other agents have the most significant influence on a certain agent, we employ a double-attention mechanism on both topological and temporal features.

Besides models that directly maximize the data likelihood, deep generative models such as generative adversarial networks (GAN), variational auto-encoder (VAE) and their conditional variants have been widely applied to representation learning and distribution approximation tasks \cite{goodfellow2014generative, kingma2013auto}. 
Despite that VAE is a highly flexible latent variable model with encoder-decoder architecture which makes the posterior of the latent variable as similar as its prior (usually a normal distribution), the two distributions do not match well in many tasks, which breaks the consistency of the model. 
Also, although GAN have achieved satisfying performance on image generation tasks, it usually suffers from mode collapse problems especially when applied to sequential data under the conditional setting. Some variants of VAE \cite{ma2019,zhao2017infovae} have been developed to mitigate these drawbacks by incorporating information theory, which are similar to the generative component in this work.

The main contributions of this paper are summarized as:

$\bullet$ We propose a generic trajectory forecasting system with relational reasoning between interactive agents to predict pedestrian and vehicle trajectories.

$\bullet$ We design a graph double-attention network (GDAT) to extract and update node features for spatio-temporal dynamic graphs with a novel topological attention mechanism based on a kernel function. The main advantages of GDAT are three-fold: (a) It can be applied to flexible number of agents with the property of permutation invariance, which enhances generalization ability; (b) It can model high-order interactions by multiple loops of message passing; (c) The attention mechanism provides a heuristic for multi-agent interaction modeling.

$\bullet$ We incorporate an efficient kinematic constraint layer to ensure physical feasibility for vehicle trajectory prediction, which can also smooth the trajectories and reduce prediction error.

$\bullet$ We validate the proposed system on multiple trajectory forecasting benchmarks and the system achieves the state-of-the-art performance.

The remainder of the paper is organized as follows.
Section 2 provides a brief overview on state-of-the-art related research.
Section 3 presents a generic problem formulation for trajectory prediction tasks. 
Section 4 illustrates the proposed forecasting system Social-WaGDAT. In Section 5, the proposed system is applied to interactive pedestrian and vehicle trajectory prediction based on real-world benchmark datasets. The performance is compared with baseline methods in terms of widely-used evaluation metrics. 
Finally, Section 6 concludes the paper.

\begin{figure*}[!tbp]
    \centering
    \includegraphics[width=\textwidth]{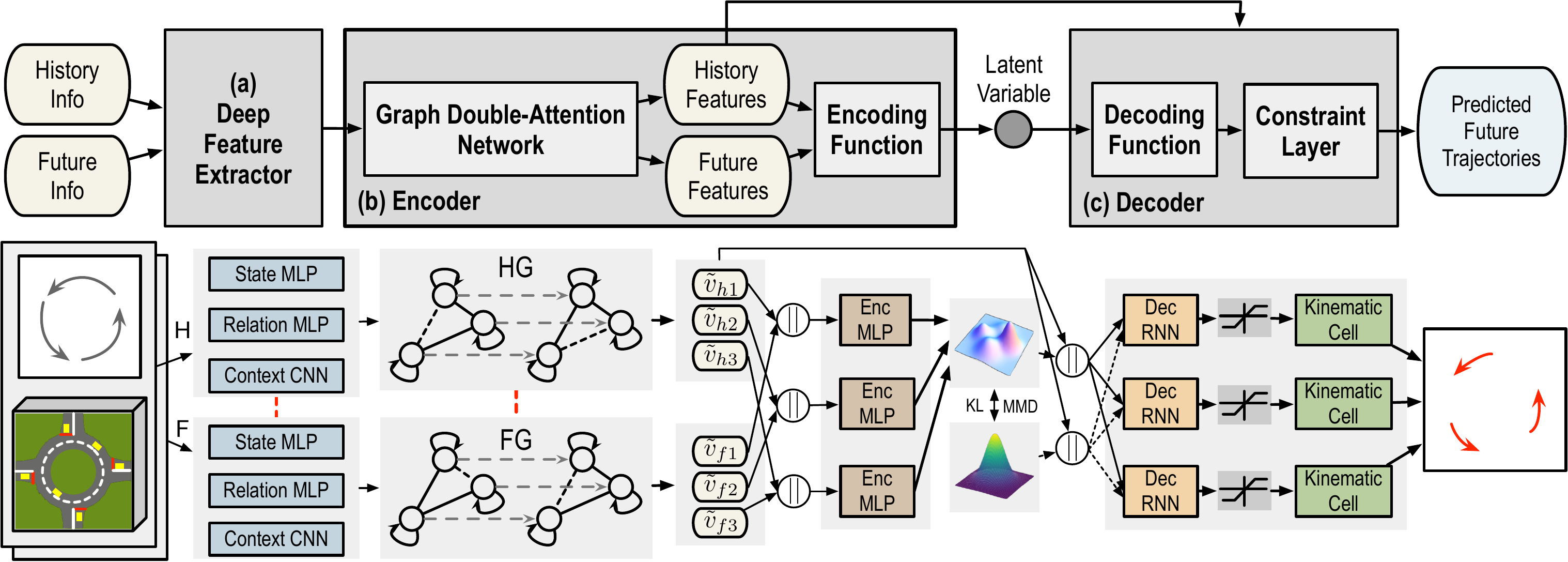}
    \caption{ The detailed architecture of Social-WaGDAT, which consists of three key components: (a) A deep feature extractor which extracts state, relation and context features from the trajectories of agents, the sequences of occupancy density maps and velocity fields. The red dashed lines indicate sharing parameters.
    (b) An encoder which includes a graph double-attention network that processes spatio-temporal graphs and generates abstract node attributes containing interaction information, and an encoding function which maps the node attributes to a latent space. During the testing phase, the encoding function is not used.
    (c) A decoder which samples future trajectory hypotheses satisfying physical constraints for each agent. 
    The bottom portion of the figure presents some details of (a)-(c).
    $||$ denotes the concatenation operation. MLP refers to multi-layer perceptron.}
    \vspace{-0.3cm}
    \label{fig:model}
\end{figure*} 

\section{Related Work}
In this section, we provide a brief literature review on related research and illustrate the distinction and advantages of the proposed generative trajectory prediction framework.

\vspace{0.05cm}
\noindent\textbf{Trajectory and Behavior Prediction}

Extensive research has been conducted on trajectory prediction for pedestrians and on-road vehicles. Early literature mainly introduced physics-based or rule-based approaches such as state estimation techniques applied to kinematic models (e.g. constant velocity model), which can only perform well for short-term prediction with very limited model capacity \cite{liu2016vehicle,scharcanski2010particle}. As machine learning techniques are studied more extensively, people began to employ learning-based models such as hidden Markov models \cite{wang2018learning}, Gaussian mixture models \cite{li2018generic,Jiachen_ITSC18-2}, dynamic Bayesian network \cite{kasper2012object} and inverse reinforcement learning \cite{sun2018probabilistic}. 
In recent years, many deep learning models have been proposed, which enables more flexibility and capacity to capture underlying interactive behavior patterns \cite{fernando2018soft+,lee2017desire,rudenko2019human,xu2018encoding,liang2019peeking,ma2019trafficpredict,Jiachen_IROS19,li2018development,huang2019uncertainty,su2019potential}. 
However, physics-based feasibility constraints are usually ignored in learning-based methods.
In this paper, we address interaction modeling and introduce a probabilistic trajectory prediction system based on deep generative models with interpretation from information theory, which also takes feasibility constraints into consideration.

\vspace{0.05cm}
\noindent\textbf{Relational Reasoning and Graph Networks}

The goal of relational reasoning is to figure out the relationships among different entities, such as image pixels \cite{wang2018non}, words or sentences \cite{lin2017structured}, human skeletons \cite{battaglia2016interaction} or interactive moving agents \cite{zambaldi2018relational,choi2019looking}. A typical representation of the whole context is to formulate a graph, where object states are node features and their relationships are edge features. Recently, graph networks (GN) are employed widely as a tool of graph-based learning, where there is no restriction on the message passing rules. Most existing works focused on the approximation function parameterized by deep neural network due to its high flexibility, which leads to graph neural networks (GNN). In this paper, we design a graph network with both topological and temporal attention mechanisms to capture underlying patterns of situation evolution.

\vspace{0.05cm}
\noindent\textbf{Deep Generative Modeling}

One of the advantages of generative modeling lies in the data distribution learning without supervision. Coupled with highly flexible deep networks, deep generative models have achieved satisfying performance in image generation, style transfer, sequence synthesis tasks, etc. Besides GAN and VAE, the Wasserstein auto-encoder (WAE) \cite{tolstikhin2017wasserstein} was proposed from the optimal transport point of view, which encourages the consistency between the encoded latent distribution and the prior. Also, \cite{zhao2017infovae} proposed a variant of VAE combined with information theory and a similar approach was proposed in \cite{ma2019}.

\section{Problem Formulation}
The goal of this work is to predict future trajectories for multiple interactive agents based on their historical states and context information, where the number of involved agents may vary in different cases.
Without loss of generality, we assume there are $N$ agents in the scene. We denote a set of trajectories covering the historical and forecasting horizons ($T_h$ and $T_f$) as
$\mathbf{T}_{1:T}=\{\tau^i_{1:T}|\tau^i_{k}=(x^i_k, y^i_k,\psi^i_k),T=T_h+T_f,i=1,...,N\}$,
where $(x,y)$ is the 2D coordinate in the world space or pixel space and $\psi$ is the heading angle, which is not necessary for pedestrian prediction. 
We also denote a sequence of context information (images or tensors) as $\mathbf{C}_{1:T}=\{c_{1:T},T=T_h+T_f\}$ for global context or $\mathbf{C}_{1:T}=\{c^i_{1:T}, T=T_h+T_f, i=1,...,N\}$ for the local context of each agent.
The future information is accessible during training.
The trajectory and context information can be transformed to arbitrary format within the model as long as the output is consistent.
Then we aim to approximate the conditional distribution $p(\mathbf{T}_{T_h+1:T_h+T_f} | \mathbf{T}_{1:T_h}, \mathbf{C}_{1:T_h})$ with the proposed method.
Note that we only deal with fixed scenes in this work.

\section{Social-WaGDAT}
In this section, we first provide an overview of the key modules and the architecture of the proposed generative trajectory prediction system. The detailed theories and model design of each module will then be further illustrated.

\subsection{System Overview}
The detailed architecture of Social-WaGDAT is shown in Figure \ref{fig:model}, where a standard encoder-decoder architecture is employed.
There are three key components: a deep feature extractor, an encoder with graph generation and processing module and a decoder with kinematic constraint layer.
First, the feature extractor outputs state, relation and context feature embeddings from both history and future information, which includes the trajectories of the involved interactive agents and a sequence of context density maps and mean velocity fields.
The extracted features are utilized to generate a spatio-temporal graph for both the history and the future respectively, in which the node attributes are updated with a double-attention mechanism.
Then the new node attributes are transformed from the feature space into a latent space by an encoding function. Finally, the decoder generates feasible and realistic future trajectories for all the involved agents.
The number of agents can be flexible in different cases due to the weight sharing and permutation invariance of graph representation.
All the components are implemented with deep neural networks thus can be trained end-to-end efficiently and consistently.

\subsection{Feature Extraction}
The feature extractor consists of three parts: State MLP, Relation MLP and Context CNN. 
The State MLP embeds the position, velocity and heading information into
a state feature vector for each agent.
The Relation MLP embeds the relative information between each pair of agents into a relation feature vector, where the order of the two involved agents does matter since different orders correspond to different edges in the downstream graph representation. The relative information can be either the distance and relative angle (in a 2D polar coordinate) or the differences between the positions of the two agents along two axes (in a 2D Cartesian coordinate).
The Context CNN extracts spatial features for each agent from a local occupancy density map ($H\times W \times 1$) as well as heuristic features from a local velocity field ($H\times W \times 2$) centered on the corresponding agent. The reason of using occupancy density maps instead of real scene images is to remove redundant information and efficiently represent data-driven drivable regions.
The above procedures are applied at each time step, which generate a sequence of state, relation and context feature embeddings, respectively.

\subsection{Encoder with GDAT}
After obtaining the extracted features, a history graph (HG) and a future graph (FG) are generated respectively to represent the information related to the involved agents, where the state features and context features are concatenated to be the node attributes and the relation features are used as edge attributes. Note that the edge attributes are different for the same edge with different directions, which encode spatial relationship. The HG and FG are generated and processed in a similar fashion but with different time stamps.
The number of nodes (agents) in a graph is assumed to be fixed, but the edges are eliminated if the spatial distance between two nodes is larger than a threshold. Therefore, the graph topology at different time steps may vary but not influence the following procedures.

The proposed graph double-attention network consists of two consecutive layers: a \textit{topological attention layer} which updates node attributes from the spatial or topological perspective, and a \textit{temporal attention layer} which outputs a high-level feature embedding for each node, which summaries both the topological and temporal information.
Assume there are totally $n$ nodes (agents) in a graph, We denote a graph as $\mathcal{G}=\{\mathcal{V},\mathcal{E}\}$, where $\mathcal{V}=\{v_i \in \mathbb{R}^{D_n}, i\in \{1,...,n\}\}$ and $\mathcal{E}=\{e_{ij} \in \mathbb{R}^{D_e}, i,j\in \{1,...,n\}\}$. $D_n$ and $D_e$ are the dimensions of node attributes and edge attributes. 

\noindent\textbf{Topological Attention Layer}

The inputs of this layer are the original generated graphs and the output is a new set of node attributes $\Bar{\mathcal{V}}=\{\bar{v}^t_i \in \mathbb{R}^{\bar{D}_n}, i\in \{1,...,n\}, t\in \{1,...,T\}, T=T_h+T_f\}$ which can capture local structural properties. 
The topological attention coefficients $\alpha_{ij}$ (showing the significance of node $j$ w.r.t. node $i$) are calculated by
\begin{equation}
\small
    \alpha_{ij} = \frac{\exp{(-A_{ij}(\lambda\norm{v_i-v_j}^2+\mu\norm{e_{ij}}^2))}}{\sum_{k\in N(i)}\exp{(-A_{ik}(\lambda\norm{v_i-v_k}^2+\mu\norm{e_{ik}}^2))}},
\end{equation}
where $N(i)$ is the neighbor nodes with direct edges to node $i$, 
$A_{ij}$ is a prior attention coefficient which provides inductive bias from prior knowledge, $\lambda$ and $\mu$ are weight parameters to adjust the relative significance of node attributes and edge attributes when computing attention coefficients.
The intuition is that the agents with similar node attributes to the objective agent or with small spatial distance should be paid more attention to.
In this work, we set $A_{ij}=1$ implying no prior attention bias while more exploration on incorporating prior knowledge is left for future work.
Then the node attributes are updated by
\begin{equation}
\small
    \bar{v}_i = \sum_{j\in N(i)} f_{\text{act}}(\alpha_{ij}W_n v_j).
\end{equation}
The procedures of node attribute update are applied to each time step and the weight matrices and vectors are shared across different time steps.
We also employ the multi-head attention mechanism \cite{velivckovic2017graph} to boost model performance by adjusting $\lambda$ and $\mu$, where the node attributes obtained by different attention coefficients are concatenated into a whole vector.
The above message passing procedures can be applied multiple times to capture high-order interactions.

\noindent\textbf{Temporal Attention Layer}

The input of this layer is the output of the topological attention layer, which is a set of node attributes $\Bar{\mathcal{V}}=\{\bar{v}^t_i \in \mathbb{R}^{\bar{D}_n}, i\in \{1,...,n\}, t\in \{1,...,T\}\}$.
The output is a set of highly abstract node attributes $\widetilde{\mathcal{V}}=\{\widetilde{v}_i \in \mathbb{R}^{\widetilde{D}_n}, i\in \{1,...,n\}\}$ which will be further processed by the downstream modules.
The temporal attention coefficients $\beta^t_i$ is
\begin{equation}
\small
\begin{aligned}
    \beta^{ht}_i =& \frac{\exp{(f_{\text{act}}(\bar{v}^{t \top}_i w))}}{\sum^{T_h}_{t'=1} \exp{(f_{\text{act}}(\bar{v}^{t'}_i w))}}, \\
    \beta^{ft}_i =& \frac{\exp{(f_{\text{act}}(\bar{v}^{t \top}_i w)^t)}}{\sum^{T}_{t'=T_h+1} \exp{(f_{\text{act}}(\bar{v}^{t'}_i w))}},
\end{aligned}
\end{equation}
where $w \in \mathbb{R}^{\bar{D}_n}$ is a weight vector parameterizing the attention function.
Then the node attributes are updated by
\begin{equation}
\small
    \widetilde{v}^h_i = \sum^{T_h}_{t=1} f_{\text{act}}(\beta^t_i \bar{v}^{t \top}_i w), \quad
    \widetilde{v}^f_i = \sum^{T}_{t=T_h+1} f_{\text{act}}(\beta^t_i \bar{v}^{t \top}_i w).
\end{equation}
The multi-head attention mechanism can also be employed by using multiple different $w$ with average or concatenation.

\vspace{0.1cm}
\noindent
\textbf{Feature Encoding}

The historical and future node attributes are concatenated and transformed by an encoding function $f_{enc}$ to obtain the latent variable $z_i$, which is given by
\begin{equation}
    z_i = f_{enc}([\widetilde{v}^h_i||\widetilde{v}^f_i]).
\end{equation}

\subsection{Decoder with Kinematic Constraint}
\begin{figure}[!tbp]
    \centering
    \includegraphics[width=\columnwidth]{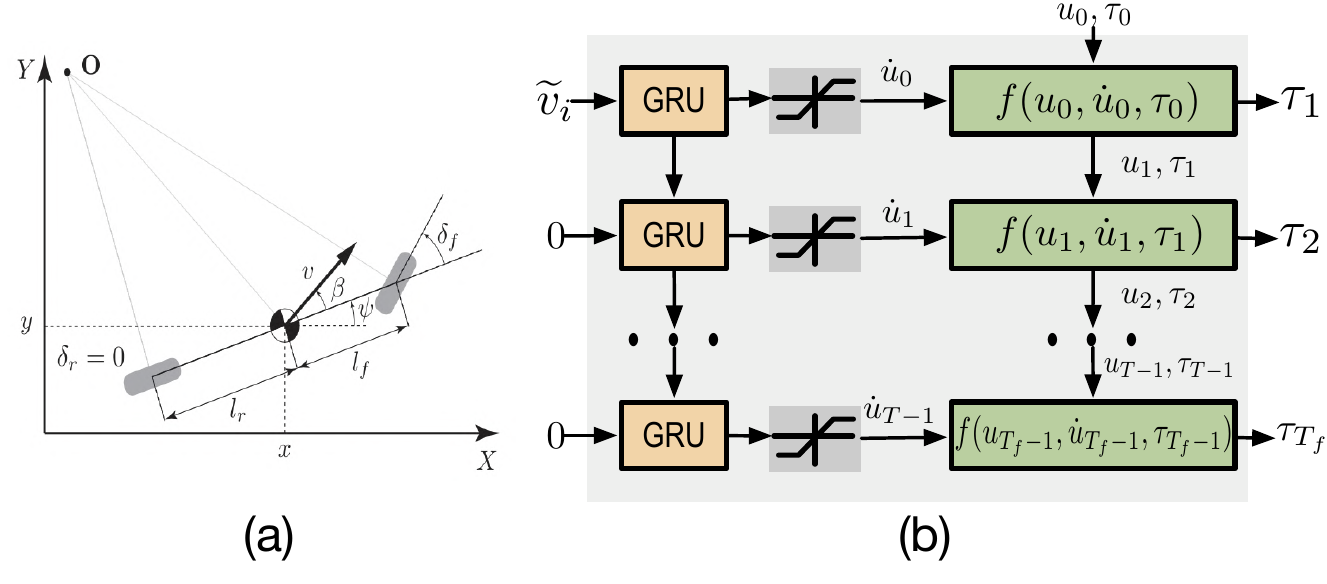}
    \caption{(a) The diagram of the kinematic bicycle model \cite{kong2015kinematic}. (b) The detailed unrolled recurrent structure of decoder for a single agent, which consists of a GRU, a saturation function and a kinematics update cell.}
    \label{fig:bicycle_model}
    \vspace{-0.2cm}
\end{figure}

We employ a similar method in \cite{ma2019} to impose a kinematic constraint cell to enforce feasible trajectory prediction following the recurrent unit, which is shown in Figure. \ref{fig:bicycle_model}.
The bicycle model is a widely used nonlinear model to approximate the kinematics of vehicles, which has a continuous form
\begin{equation}
\label{KC_equ1}
\begin{aligned}
\left \{
    \begin{aligned}
    &\dot{x} = v\cos{(\psi+\beta)}\\
    &\dot{y} = v\sin(\psi+\beta)\\
    &\dot{\psi} = \frac{v}{l_r}\sin(\beta)\\
    \end{aligned}
\right.
\end{aligned}
\end{equation}
where $x,y$ are the coordinates of the center of mass, $\psi$ is the inertial heading and $v$ is the speed of the vehicle. 
$\beta$ is the angle of the current velocity of the center of mass with respect to the longitudinal axis of the car. 
We denote $\tau_t=[x_t \ y_t \ \psi_t]^\top$, $a_t=[\dot{x}_t \ \dot{y}_t \ \dot{\psi}_t]^\top$ and $u_t=[v_t \ \beta_t]^\top$. Then the discretized version of the equations (\ref{KC_equ1}) can be written as 
\begin{equation}
\begin{aligned}
\left \{
\begin{aligned}
\tau_{t+1} &= \tau_{t} + a_t \Delta t \\
u_{t+1} &= u_t + \dot{u}_t \Delta t \\
a_t &= f(u_t, \dot{u}_t, \tau_t)\\
\end{aligned}
\right.
\end{aligned}
\end{equation}
Here we provide an instance for the agent $i$. The inputs of the gated recurrent unit (GRU) are the node attribute $\widetilde{v}_i$ at the first step and zero paddings for the following steps. The outputs are the raw $\dot{u}_t$ at each step which are truncated by a saturation function in order to restrict the accelerations in the feasible ranges. Then the kinematic cell takes in a sequence of $u,\dot{u},s$ and outputs future trajectories.
If $l_r$ is not a prior knowledge or cannot be observed, then we can either approximate it with a constant or make it an independent output.

\subsection{Loss Function and Training}
In this part, we put forward a variant of generative modeling method, which is formulated similar in \cite{ma2019} and \cite{zhao2017infovae}. In order to keep consistent with other literature on generative models, we use the same notations: $x$ denotes the predicted trajectories, $z$ denotes the latent variable and $y$ denotes the condition variable which is historical features. These notations were used for other purposes in previous sections.

The optimization problem can be written as
\begin{equation}
\begin{aligned}
\min_{\theta,\phi, s.t. 0<1-\alpha<\beta} &-\mathbf{E}_{p_{\phi}(z|x, y)}[\log p_{\theta}(x|z, y)]]\\
&+\alpha\mathbf{E}_{p(x|y)}[D_{\text{KL}}[p_{\phi}(z|x, y)||p(z|y)]]\\
&+\beta D(p_{\phi}(z|y), p(z|y)).\\
\end{aligned}
\end{equation}

Since the whole system is fully differentiable, we can train the network end-to-end by the Adam optimizer \cite{kingma2014adam} and the loss function is given by
\begin{equation}
    \small
    \begin{aligned}
    \mathcal{L} =& \gamma \ \mathbf{E}_{i \in \{1,...,N_b\}} \norm{\tau^i_{T_h+1:T} - \hat{\tau}^{i}_{T_h+1:T}}_2 \\ 
    &+ \alpha \ \mathbf{E}_{p(x|y)}[D_{\text{KL}}[p_{\phi}(z|x,y)||p(z|y)] \\
    &+ \beta \ \text{MMD}(p_{\phi}(z|y), p(z|y)),
    \end{aligned}
\end{equation}
where $\gamma$ is a weight parameter to adjust the relative importance of the reconstruction loss and $N_b$ is the total number of training agents, $D_{KL}$ is Kullback-Leibler divergence and MMD is maximum mean discrepancy. If $\gamma \gg \alpha,\beta$, then the loss function degenerates to the pure $l_2$-norm loss.

\begin{table*}[!tbp]
	\caption{ADE / FDE (meters) Comparisons of Pedestrian Trajectory Prediction (ETH \& UCY datasets). }
	\vspace{-0.3cm}
	\fontsize{8}{8}\selectfont
	\resizebox{\textwidth}{!}{
    		\begin{tabular}{m{1.5cm}<{\centering}| m{1.5cm}<{\centering}| m{1.5cm}<{\centering}| m{1.5cm}<{\centering}| m{1.5cm}<{\centering}| m{1.5cm}<{\centering}| m{1.5cm}<{\centering}| m{1.5cm}<{\centering}| m{1.5cm}<{\centering} }
    			\toprule
    			\midrule
    		    Scenes &  LR & P-LSTM & S-LSTM & S-GAN & S-GAN-P & S-Ways & SoPhie & \textbf{Social-WaGDAT}\\ 
    			\midrule 
    			ETH   &  1.33 / 2.94  &  1.13 / 2.38  &  1.09 / 2.35  &  0.81 / 1.52  &  0.87 / 1.62  &  \textbf{0.39} / \textbf{0.64}  &  0.70 / 1.43  &  0.52 / 0.91 \\ 
    			HOTEL &  \textbf{0.39} / 0.72  &  0.91 / 1.89  &  0.79 / 1.76  &  0.72 / 0.61  &  0.67 / 1.37  &  \textbf{0.39} / \textbf{0.66}  &  0.76 / 1.67  &  0.61 / 0.87 \\ 
    			UNIV  &  0.82 / 1.59  &  0.63 / 1.36  &  0.67 / 1.40  &  0.60 / 1.26  &  0.76 / 1.52  &  0.55 / 1.31  &  0.54 / 1.24  &  \textbf{0.43} / \textbf{1.12} \\ 
    			ZARA1 &  0.62 / 1.21  &  0.44 / 0.84  &  0.47 / 1.00  &  0.34 / 0.69  &  0.35 / 0.68  &  0.44 / 0.64  &  \textbf{0.30} / 0.63  &  0.33 / \textbf{0.62} \\ 
    			ZARA2 &  0.77 / 1.48  &  0.51 / 1.16  &  0.56 / 1.17  &  0.42 / 0.84  &  0.42 / 0.84  &  0.51 / 0.92  &  0.38 / 0.78  &  \textbf{0.32} / \textbf{0.70} \\ 
    			\midrule
    			AVG   &  0.79 / 1.59  &  0.72 / 1.53  &  0.72 / 1.54  &  0.58 / 1.18  &  0.61 / 1.21  &  0.46 / \textbf{0.84}  &  0.54 / 1.15   & \textbf{0.44} / \textbf{0.84} \\ 
    			\bottomrule
    		\end{tabular}
	}
	\label{tab:ETH_UCY}
	\vspace{-0.1cm}
\end{table*}

\begin{table*}[!tbp]
	\caption{ADE / FDE (pixels) Comparisons of Pedestrian Trajectory Prediction (SDD dataset).}
	\vspace{-0.3cm}
	\fontsize{8}{8}\selectfont
	\resizebox{\textwidth}{!}{
    		\begin{tabular}{m{1.5cm}<{\centering}| m{1.5cm}<{\centering}| m{1.5cm}<{\centering}| m{1.5cm}<{\centering}| m{1.5cm}<{\centering}| m{1.5cm}<{\centering}| m{1.5cm}<{\centering}| m{1.5cm}<{\centering}| m{1.5cm}<{\centering} }
    			\toprule
    			\midrule
    		     &  P-LSTM  & S-Forces & S-LSTM & S-GAN & CAR-Net & S-ATT & DESIRE & \textbf{Social-WaGDAT}\\ 
    			\midrule 
    			  &  38.12 / 58.63 & 36.48 / 58.14  & 33.19 / 56.38  & 28.67 / 44.35 & 25.72 / 51.80 & 33.28 / 55.86 &  35.38 / 57.62 & \textbf{22.52} / \textbf{38.27} \\ 
    			\bottomrule
    		\end{tabular}
	}
	\label{tab:SDD}
	\vspace{-0.1cm}
\end{table*}

\begin{table*}[!tbp]
	\caption{ADE / FDE (meters) Comparisons of Vehicle Trajectory Prediction (ID dataset).}
	\vspace{-0.3cm}
	\fontsize{8}{8}\selectfont
	\resizebox{\textwidth}{!}{
    		\begin{tabular}{m{0.8cm}<{\centering}|m{0.6cm}<{\centering}|m{1.3cm}<{\centering}| m{1.3cm}<{\centering}| m{1.3cm}<{\centering}| m{1.3cm}<{\centering}| m{1.3cm}<{\centering}| m{1.5cm}<{\centering}| m{1.5cm}<{\centering}| m{1.5cm}<{\centering}| m{1.5cm}<{\centering}}
    			\toprule
    			\midrule
    			\multirow{4}*{\shortstack[lb]{}} 
    		 &	& \multicolumn{5}{c|}{Baseline Methods} & \multicolumn{4}{c}{\textbf{Social-WaGDAT}} \\
    			\cline{3-11}
    			 & & & & & & & & & &\\[-0.1cm]
    		     Scenes & Time  & CVM & P-LSTM & S-LSTM & S-GAN & S-ATT & $\mathbf{T}$  & $\mathbf{T}+\mathbf{C}+\textbf{\text{w/o Attention}}$ & $\mathbf{T}+\mathbf{C}$ &  $\mathbf{T}+\mathbf{C}+\textbf{\text{Kinematic}}$\\[-0.1cm]
    		     & & & & & & & & & &\\
    			\midrule 
    			\multirow{5}*{RA}  
		&1.0s	& 0.79 / 1.04 &  0.70 / 0.94  &   0.44 / 0.62  &  0.34 / 0.48  &  0.32 / 0.49  & 0.27 / 0.41  & 0.31 / 0.45  & 0.24 / 0.36  & \textbf{0.24} / \textbf{0.35}  \\ 
		&2.0s	& 1.54 / 2.73 &  1.24 / 2.04  &   0.93 / 1.66  &  0.69 / 1.24  &  0.84 / 1.72  & 0.67 / 1.22 & 0.76 / 1.26 & 0.63 / 1.15 & \textbf{0.60} / \textbf{1.01} \\ 
		&3.0s	& 2.27 / 3.91 &  1.75 / 2.98  &   1.44 / 2.71  &  1.10 / 2.15  &  1.51 / 3.15  & 1.01 / 2.11 & 1.19 / 2.34 & 0.94 / 2.04 & \textbf{0.86} / \textbf{1.83}  \\ 
		&4.0s	& 2.73 / 4.02 &  2.09 / 4.21  &   1.86 / 4.27  &  1.56 / 3.25  &  1.93 / 3.50  & 1.34 / 2.84 & 1.51 / 2.89 & 1.26 / 2.55  & \textbf{1.08} / \textbf{2.21} \\ 
		&5.0s   & 2.90 / 4.38 &  2.35 / 5,39  &   2.17 / 5.12  &  2.11 / 4.70  &  2.13 / 4.86  & 1.85 / 4.41 & 2.13 / 4.24 & 1.65 / 3.98 & \textbf{1.31} / \textbf{3.34}  \\ 
		    \midrule
		    \multirow{5}*{UI}  
		&1.0s	& 0.88 / 1.14 &  0.74 / 1.00  &   0.50 / 0.70  &  0.41 / 0.57  &  0.38 / 0.57  & 0.40 / 0.59 & 0.41 / 0.61 & 0.37 / 0.55 & \textbf{0.36} / \textbf{0.54} \\ 
		&2.0s	& 1.64 / 2.85 &  1.30 / 2.14  &   1.00 / 1.79  &  0.82 / 1.49  &  0.99 / 2.04  & 0.81 / 1.41 & 0.87 / 1.56 & 0.78 / 1.38 & \textbf{0.74} / \textbf{1.31}  \\ 
		&3.0s	& 2.44 / 4.33 &  1.88 / 3.34  &   1.63 / 3.26  &  1.34 / 2.70  &  1.79 / 3.71  & 1.21 / 2.46 & 1.25 / 2.32 & 1.15 / 2.12 & \textbf{1.07} / \textbf{2.04} \\ 
		&4.0s	& 2.91 / 4.27 &  2.39 / 4.05  &   2.22 / 4.19  &  1.95 / 4.16  &  2.43 / 4.50  & 1.86 / 3.98 & 1.93 / 4.26  & 1.78 / 3.80 & \textbf{1.63} / \textbf{3.56} \\ 
		&5.0s   & 3.14 / 5.02 &  2.77 / 4.33  &   2.69 / 4.64  &  2.65 / 4.87  &  2.85 / 4.90  & 2.52 / 4.57 & 2.44 / 4.51 & 2.27 / 3.90 & \textbf{1.99} / \textbf{3.85} \\ 
		    \midrule
		    \multirow{5}*{HR}  
		&1.0s	& \textbf{0.33} / \textbf{0.50} &  0.39 / 0.54  &   0.36 / 0.52  &  0.40 / 0.53  &  0.38 / 0.53  & 0.42 / 0.51 & 0.44 / 0.56  & 0.40 / 0.54 & 0.40 / 0.53  \\ 
		&2.0s	& 0.81 / 1.30 &  0.80 / 1.41  &   0.80 / 1.49  &  0.69 / \textbf{1.13}  &  0.72 / 1.43  & 0.71 / 1.26 & 0.73 / 1.49  & 0.68 / 1.31 & \textbf{0.67} / 1.28 \\ 
		&3.0s	& 1.15 / 2.00 &  1.23 / 2.28  &   1.29 / 2.49  &  1.00 / 1.81  &  1.19 / 2.31  & 0.91 / 1.75 & 0.98 / 1.97 & 0.88 / 1.78 &  \textbf{0.87} / \textbf{1.78} \\ 
		&4.0s	& 1.45 / 2.45 &  1.64 / 3.04  &   1.79 / 3.49  &  1.34 / 2.51  &  1.59 / 2.94  & 1.33 / 2.44 & 1.58 / 2.78 & 1.34 / 2.40  & \textbf{1.31} / \textbf{2.31} \\ 
		&5.0s   & 1.71 / 3.25 &  1.95 / 3.26  &   2.19 / 3.87  &  1.68 / 3.22  &  1.89 / 3.04  & 1.66 / 3.13 & 1.81 / 3.34 & 1.62 / 2.97 &  \textbf{1.57} / \textbf{2.88} \\ 
			\bottomrule
    		\end{tabular}
	}
	\label{tab:ID}
	\vspace{-0.cm}
\end{table*}

\section{Experiments}
In this section, we validate the proposed method on three publicly available benchmark datasets for trajectory prediction of pedestrians and on-road vehicles. The results are analyzed and compared with state-of-the-art baselines. 

\subsection{Datasets}
Here we briefly introduce the datasets below. Please refer to the supplementary materials for details about the data processing procedures.

\vspace{0.1cm}
\noindent
\textbf{ETH} \cite{ETH} \textbf{and UCY} \cite{UCY}: 
These two datasets are usually used together in literature, which include top-down-view videos and image annotations of pedestrians in both indoor and outdoor scenarios. The trajectories were extracted in the world space.

\vspace{0.1cm}
\noindent
\textbf{Stanford Drone Dataset (SDD)}\cite{SDD}:
The dataset also contains a set of top-down-view images and the corresponding trajectories of involved entities, which was collected in multiple scenarios in a university campus full of interactive pedestrians, cyclists and vehicles. The trajectories were extracted in the image pixel space.

\vspace{0.1cm}
\noindent
\textbf{INTERACTION Dataset (ID)}\cite{interactiondataset}:
The dataset contains naturalistic motions of various traffic participants in a variety of highly interactive driving scenarios. Trajectory data was collected using drones and traffic cameras. 

The semantic maps of scenarios and agents' trajectories are provided. We consider three types of scenarios: roundabout (RA), unsignalized intersection (UI) and highway ramp (HR). The trajectories were extracted in the world space.

\subsection{Evaluation Metrics and Baselines}
We evaluate the model performance in terms of average displacement error (ADE) defined as the average distance between the predicted trajectories and the ground truth over all the involved entities within the prediction horizon, as well as final displacement error (FDE) defined as the deviated distance at the last predicted time step. 

For the ETH, UCY and SDD dataset, 
we predicted the future 12 time steps (4.8s) based on the historical 8 time steps (3.2s).
For the ID dataset, we predicted the future 10 time steps (5.0s) based on the historical 4 time steps (2.0s).

We compared the performance of our proposed method with the following baseline approaches: Constant Velocity Model (CVM), Linear Regression (LR), Probabilistic LSTM (P-LSTM) \cite{Jiachen_ICRA19}, Social Forces (S-Forces) \cite{luber2010people}, Social LSTM (S-LSTM) \cite{alahi2016social}, Social GAN (S-GAN and S-GAN-P) \cite{gupta2018social}, Social Attention (S-ATT) \cite{vemula2018social}, Social Ways (S-Ways) \cite{amirian2019social}, SoPhie \cite{sadeghian2019sophie}, CAR-Net \cite{sadeghian2018car} and DESIRE \cite{lee2017desire}. Please refer to the reference papers for more details.

\subsection{Implementation Details}
A batch size of 64 was used and the models were trained for 100 epochs using Adam with an initial learning rate of 0.001. The models were trained on a single TITAN X GPU. We used a split of 70\%, 10\%, 20\% as training, validation and testing data.
Please refer to supplementary materials for more details on the model architecture.

\subsection{Quantitative Analysis}
\textbf{ETH and UCY Datasets}:
The comparison of the proposed Social-WaGDAT and baseline methods in terms of ADE and FDE is shown in Table \ref{tab:ETH_UCY}. Some of the reported statistics are adopted from the original papers.
It is not surprising that the linear model performs the worst in general since it does not consider any social interactions or context information. An exception is the HOTEL scenario since most trajectories are relatively straight and can be well approximated by line segments.
The P-LSTM is able to achieve smaller prediction error than LR due to the larger model capacity and flexibility of recurrent neural network, although it also predicts solely based on the individual's historical trajectories.
The S-LSTM considers the interactions between entities by using a social pooling mechanism. The S-GAN and S-Ways further improve the performance by introducing deep generative modeling.
Both SoPhie and our method leverage the trajectory and context information, but in different ways. Our model can achieve better performance owing to the explicit interaction modeling with graph neural networks and more compact distribution learning with conditional Wasserstein generative modeling. 
In general, our approach achieves the smallest average ADE and FDE across different scenes. 

\textbf{Stanford Drone Dataset}:
The comparison of results is provided in Table \ref{tab:SDD}, where the ADE and FDE are reported in the pixel distance.
Note that we also included cyclists and vehicles in the test set besides pedestrians.
Similarly, the P-LSTM performs the worst due to lack of relational reasoning. The S-Forces incorporates interaction modeling from an energy-based perspective, while the S-ATT and CAR-Net utilize attention mechanisms, which leads to better accuracy.
The S-GAN and DESIRE both solve the task from a probabilistic perspective by learning implicit data distribution and latent space representations, respectively.
Our approach achieves the best performance in terms of prediction error, which implies the superiority of explicit interaction modeling and necessity of leveraging both trajectory and context information. 
The prediction error is reduced by 12.1\% with respect to the best baseline method.

\textbf{INTERACTION Dataset}:
We finally compare the model performance on the real-world driving dataset in Table \ref{tab:ID}. 
Here we only involved the baseline approaches whose codes are publicly available.
Although we trained a unified prediction model on different scenarios simultaneously, we analyzed the results for each type of scenario separately.
To allow for fair comparison, here we only compare our model $\mathbf{T}$ with baseline methods since they do not leverage context information.
In the HR scenarios, the linear model CVM has a good performance in general since most vehicles go straight along the lane without obvious velocity changes within a short period, which makes the assumption of constant velocity well applicable. But the learning-based models may be negatively affected by some subtle patterns learned from data and redundant information in such scenarios, especially for short-term prediction (e.g. 1.0s).
As the prediction horizon increases, the results of baseline methods are comparable while our model achieves the best performance. The future behaviors in HR scenarios are relatively easy to forecast so our method did not achieve a significant improvement.
In the RA and UI scenarios, however, the superiority of the proposed system is more distinguishable.
It is shown that CVM performs the worst across all forecasting horizons since turning behaviors, negotiation and interaction between vehicles happen frequently, which makes the assumption of constant velocity much less applicable. 
The P-LSTM has a slightly better performance by using the recurrent network.
While the S-LSTM, S-GAN and S-ATT incorporate interaction modeling by different strategies which further reduce the prediction error, our model $\mathbf{T}$ still performs the best. This implies the advantages of the graph representation for interaction modeling.
By using our full model $\textbf{T}+\textbf{C}+\textbf{Kinematic}$, the prediction error is reduced by 29.4\%, 21.1\% and 8.8\% in RA, UI and HR with respect to the best baseline method, respectively.

\subsection{Qualitative and Ablative Analysis}
\begin{figure*}[!tbp]
    \centering
    \includegraphics[width=\textwidth]{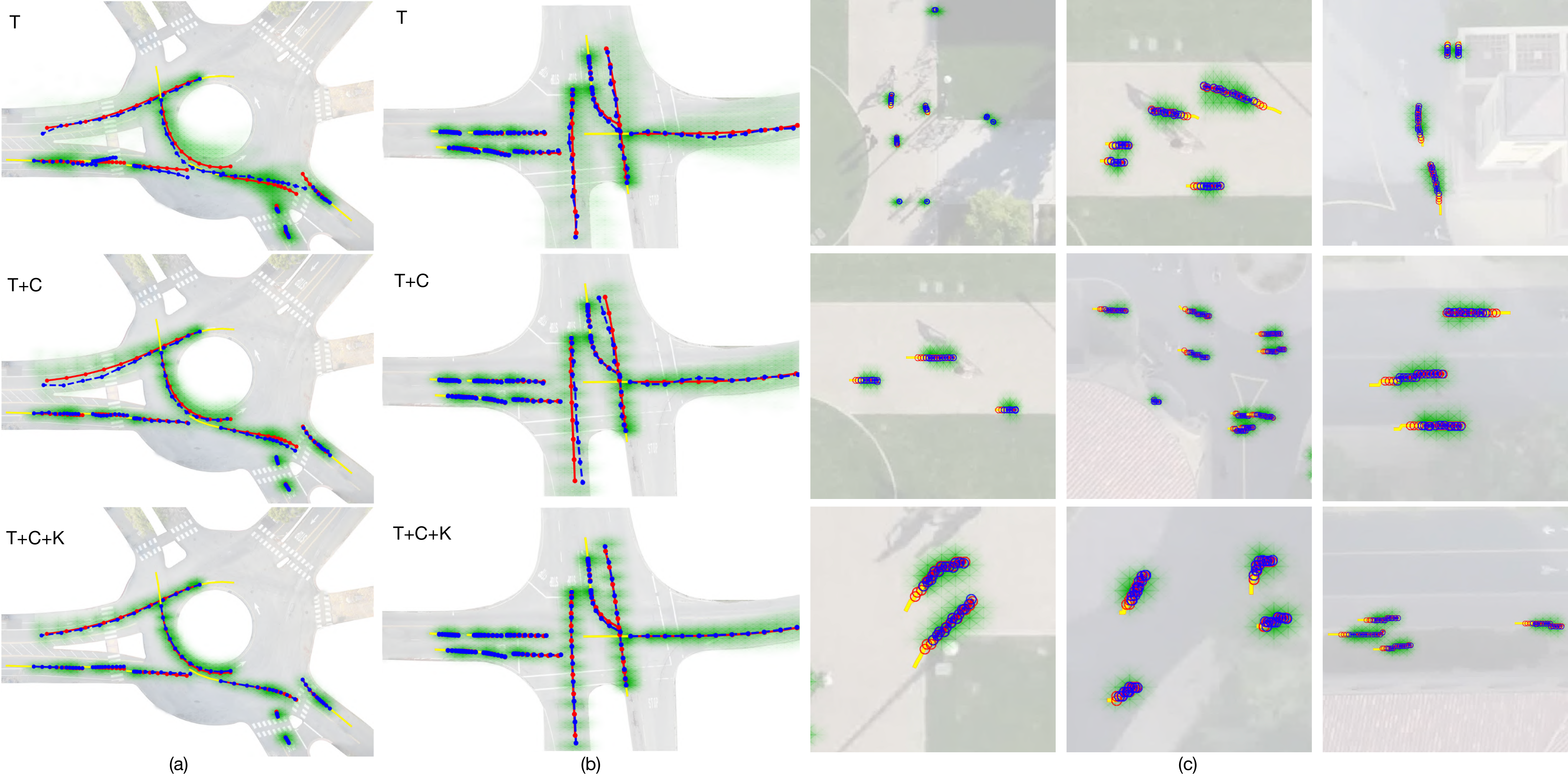}
    \caption{Qualitative and ablative results, where the green mask represents the predicted distribution and the yellow, blue and red lines represent historical observation, ground truth and a trajectory hypothesis sampled from the distribution with the smallest error, respectively. 
    (a) and (b) include images from the ID dataset with ablative demonstration and (c) includes images from the SDD dataset.
    }
    \label{fig:test_plot}
    \vspace{-0.3cm}
\end{figure*} 

We qualitatively evaluated on prediction hypotheses of typical testing cases in Figure \ref{fig:test_plot}. 
Although we jointly predict all agents in a scene, we show predictions for a subset for simplicity.
It shows that our approach can handle different challenging scenarios (e.g. intersection, roundabout) and diverse behaviors (e.g. going straight, turning, stopping) of vehicles and pedestrians. 
The ground truth is close to the mean of the predicted distribution and the model also allows for uncertainty.

We also conducted ablative analysis on the ID dataset to demonstrate relative significance of context information, double-attention mechanism and kinematic constraint layer in Social-WaGDAT. 
The ADE and FDE of each model setting are shown in the right part of Table \ref{tab:ID}.

$\bullet$ $\mathbf{T}$ versus $\mathbf{T}+\mathbf{C}$: 
We show the effectiveness of employing scene context information.
$\mathbf{T}$ is the model without the kinematic layer, which only uses trajectory information, while $\mathbf{T}+\mathbf{C}$ further employs context information. 
The models directly output the position displacements $\Delta \tau_t$ at each step, which are aggregated to get complete trajectories. 
We can see little difference on prediction errors over short horizons while the gap becomes larger as the horizon extends. The reason is that the vehicle trajectories within a short period can be approximated by the constant velocity model, which are not heavily restricted or affected by the static context. However, as the forecasting horizon increases, the effects of context constraints cannot be ignored anymore, which leads to larger performance gain of leveraging context information.
Compared with $\mathbf{T}$, the average prediction errors of $\mathbf{T}+\mathbf{C}$ are reduced by 11.1\%, 10.1\% and 2.6\% in RA, UI and HR scenarios, respectively. This implies that the context information has larger effects on the prediction in RA and UI scenarios, where the influence of road geometries cannot be ignored.
In Figure 4(a) and 4(b), the predicted distribution of $\mathbf{T}+\mathbf{C}$ is more compliant to roadways to avoid collisions and the vehicles near the ``yield'' or ``stop'' signs tend to yield or stop.
However, $\mathbf{T}$ generates samples that are outside of feasible areas and violating the traffic rules.

$\bullet$ $\mathbf{T}+\mathbf{C}+\textbf{w/o Attention}$ versus $\mathbf{T}+\mathbf{C}$: 
We show the effectiveness of the proposed double-attention mechanism.
$\mathbf{T}+\mathbf{C}+\textbf{w/o Attention}$ uses equal attention coefficients in both topological and temporal layers. 
According to the statistics reported in Table \ref{tab:ID}, compared with equal attention, employing the double-attention mechanism to figure out relative importance within the topological structure and along different time steps can reduce the average prediction error by 20.1\%, 13.9\% and 10.5\% in the RA, UI and HR scenarios, respectively.
This implies that certain agents and periods have more significant influence than the rest.

$\bullet$ $\mathbf{T}+\mathbf{C}$ versus $\mathbf{T}+\mathbf{C}+\textbf{Kinematic}$:
We show the effectiveness of the kinematic constraint layer.
Employing the kinematic constraint layer to regularize the learning-based prediction hypotheses can further reduce the average prediction error by 20.6\%, 12.3\% and 3.1\% in the RA, UI and HR scenarios, respectively.
The reason is that due to the restriction from the kinematic model, unfeasible movements can be filtered out and the model is unlikely to overfit noisy data or outliers.
Moreover, the improvement in RA and UI is more significant than in HR. The reason is that most vehicles go straight along the road in HR, whose behaviors can be well approximated by linear models. However, there are frequent turning behaviors in RA and UI which need constraints by more sophisticated models. 
We also visualize the predicted trajectories in Figure 4(a) and 4(b), where the ones with kinematic constraints are smoother and more plausible. 

\section{Conclusions}
In this paper, we propose a generic system for multi-agent trajectory prediction named Social-WaGDAT, which considers both statistical context information, trajectories of interactive agents and physical feasibility constraints. In order to effectively model the interactions between different entities, we design a graph double-attention network to extract features from spatio-temporal dynamic graphs and a topological attention mechanism to figure out relative significance. Moreover, a variant of Wasserstein generative modeling is employed to support the whole framework. The Social-WaGDAT is validated by both pedestrian and vehicle trajectory prediction tasks. The experimental results show that our approach achieves smaller prediction error than multiple baseline methods. 
For the future work, we will consider the interactions between heterogeneous agents explicitly and jointly predict their trajectories.

\newpage

{\small
\bibliographystyle{ieee_fullname}
\bibliography{egbib}
}
\newpage

\section{Model Details}
In this section, we introduce the implementation details of our model.

\begin{itemize}
    \item \textbf{Deep Feature Extractor (FE)}: The State MLP and Relation MLP both have three layers with 64 hidden units. The Context CNN consists of five layers with kernel size $5\times 5$ with zero paddings to keep the size of the image-like tensor.
    \item \textbf{Graph Double-Attention Network (GDAT)}: The dimension of node attributes are 64 and the dimension of edge attributes is 16. These dimensions are fixed in different rounds of message passing. The activation functions in the attention mechanism are LeakyReLU.
    \item \textbf{Encoding Function} (MLP): The encoding function is a three-layer MLP with 128 hidden units. The dimension of latent variable is 32.
    \item \textbf{Decoding Function}: The decoding function is a recurrent layer of GRU cells with 128 hidden units.
\end{itemize}

\section{Loss function}
In this section, we provide a supplementary introduction of the loss function we used. In \cite{zhao2017infovae}, they proposed a variation of VAE model which can maximize the mutual information between latent variable and observable variable. In our model, the original loss function is:
\begin{equation}\label{equ5}
\boxed{\begin{aligned}
\min_{\theta, \phi, s.t. 0<1-\alpha<\beta} &-\mathbf{E}_{p(x)} \mathbf{E}_{p_{\phi}(z|x)}[\log p_{\theta}(x|z)]] \\
&+\alpha\mathbf{E}_{p(x)}[D_{\text{KL}}[p_{\phi}(z|x)||p(z)] \\
&+\beta D(p_{\phi}(z), p(z))
\end{aligned}}
\end{equation}
This loss function is a Lagrange function of the following optimization problem:
\begin{equation}\label{equ6}
\begin{aligned}
&\max I_{\phi}(x, z)\\
s.t.&
\left \{
\begin{aligned}
D_{\text{KL}}&[p_{\phi}(z)||p(z)]\le \epsilon_1\\
D_{\text{KL}}&[p_{\phi}(x, z)||p_{\theta}(x, z)] \le \epsilon_2\\
\end{aligned}
\right.
\end{aligned}
\end{equation}
\begin{equation}\label{equ7}
\Leftrightarrow 
\begin{aligned}
\min_{0<1-\alpha<\beta} &-(1-\alpha) I_{\phi}(x, z) \\
&+ (\beta+ \alpha-1)\mathcal{D}_{KL}[p_{\phi}(z)||p(z)] \\
&+\mathcal{D}_{KL}[p_{\phi}(x, z)||p_{\theta}(x, z)]
\end{aligned}
\end{equation}
We assume the distribution of decoder $p_{\theta}(x|z) \sim \mathcal{N}(\mu_{\theta}(z), 1)$ and $p(z|y)$ is a standard normal distribution. We use
Maximum Mean Discrepancy (MMD) to approximate $D_{\text{KL}}[p_{\phi}(z)||p(z)]$, combining with the GNN, then we have 
\begin{equation}
    \begin{aligned}
    \mathcal{L} =& \gamma \ \mathbf{E}_{i \in \{1,...,N_b\}} \norm{\tau^i_{T_h+1:T} - \hat{\tau}^{i}_{T_h+1:T}}_2^2 \\
    +& \alpha \ \mathbf{E}_{p(x|y)}[D_{\text{KL}}[p_{\phi}(z|x,y)||p(z|y)] \\
    +& \beta \ \text{MMD}(p_{\phi}(z|y), p(z|y)),
    \end{aligned}
\end{equation}
where 
\begin{equation}
    \centering
    \begin{aligned}
    p(z|x, y) &= \mathcal{N}(MLP(GDAT(FE(x)),GDAT(FE(y))), I), \\
    \tilde{v}_h &= GDAT(FE(y)), \ \tilde{v}_f = GDAT(FE(x)),\\
    y &= \{\mathbf{T}_{1:T_h}, \mathbf{C}_{1:T_h}\}, \\
    x &= \{\mathbf{T}_{T_h+1:T}, \mathbf{C}_{T_h+1:T}\}.
    \end{aligned}
\end{equation}
$FE$ is the deep feature extractor and $GDAT$ is the proposed graph double-attention network.

\section{Data Preprocessing}
\begin{figure}[!tbp]
\begin{center}
\includegraphics[width=\columnwidth]{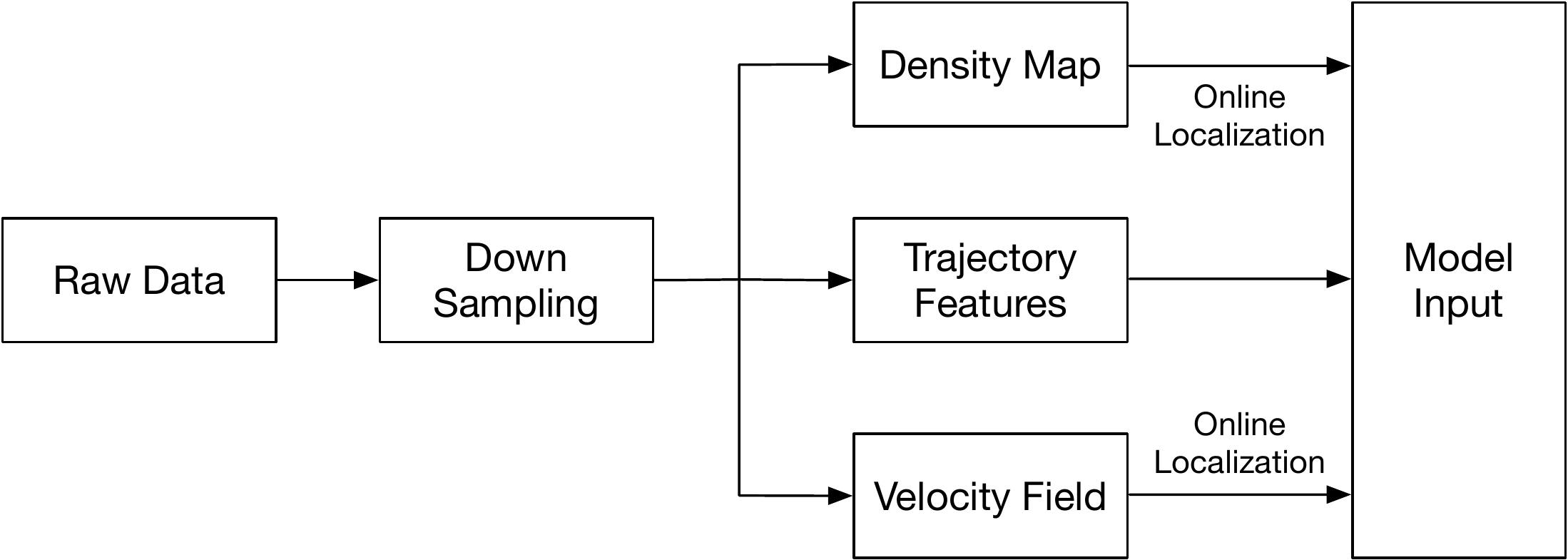}
\caption{The data preprocessing pipeline.}
\label{fig:preprocessing}
\end{center}
\end{figure}

We used a sequence of modules to preprocess the raw data, which are shown in Figure \ref{fig:preprocessing}. The details are introduced below.

\subsection{Global Context Information}
In order to provide better global context information, we designed two different representations, namely occupancy density map and mean velocity field. By constructing such global contexts offline, we did decentralized localization for the corresponding target agent and obtained their local context information. The context information was used in both training and testing phases. 
\vspace{0.1cm}

\noindent\textbf{Occupancy Density Map}
The density map describes the normalized frequency distribution of all the agents' locations. For a specific scene, we first split our map into a number of bin areas, which are $1m\times 1m$ squares. Without loss of generality, we denote this histogram as $B$, and all the agents in different frames as a set $\{o_{t,k}\}$, where $k$ is the agent index and $t$ is the frame index. We obtained the global representation of density by calculating $B_{i,j}=\sum_{t,k}\phi(o_{t,k},i,j)$, where $i,j$ are the indices of the histogram and $\phi(o_{t,k},i,j)$ is an indicator function which equals 1 if $o_{t,k}$ is located in the bin area indicated by index $i,j$ and 0 otherwise. Then we normalized this density map by dividing all bin values by the maximum value in the histogram and used this normalized histogram as our occupancy density map.
\vspace{0.1cm}

\noindent\textbf{Mean Velocity Field}
Similarly, we also created a map of velocity field which contains $1m \times 1m$ square areas. We denote the whole map as $VF$ and the bin item indexed by $i,j$ as $VF(i,j)$. The $VF(i,j)$ is a two-dimensional vector representing the average speed along vertical and horizontal axes of all the agents in this area. 
Mathematically, $VF(i,j)_{x} = \frac{1}{N}\sum_{t,k}\phi(v_{t,k},i,j)v_{t,k}^{x}$ and $VF(i,j)_{y} = \frac{1}{N}\sum_{t,k}\phi(v_{t,k},i,j)v_{t,k}^{y}$. 

\subsection{Localization}
\begin{figure}[!tbp]
\begin{center}
\includegraphics[width=0.6\columnwidth]{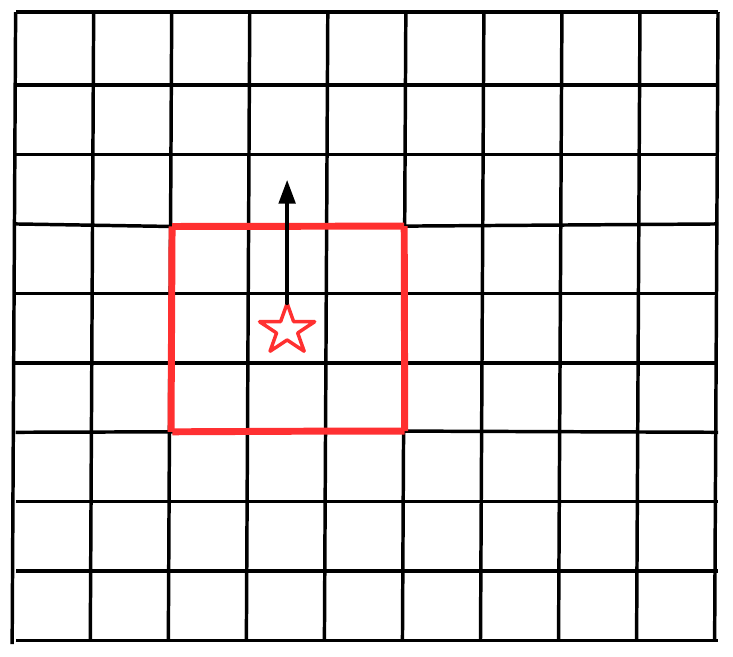}
\caption{The illustrative diagram of local context information. The target agent is denoted by the red star and its local context is the $3\times 3$ matrix denoted by the red box centered on itself.}
\label{fig:local-central}
\end{center}
\end{figure}

After obtaining the global context offline, our model utilized a decentralized method to do localization for each agent during training and testing. Given the location and the moving direction of the current agent at the current time step, we obtained the local context centered on this agent along its moving direction from the global context. Figure \ref{fig:local-central} provides an illustrative example.

\section{Baseline Methods}
In this section, we provide a more detailed introduction to the baseline methods used in our paper.
\begin{itemize}
    \item \textbf{Constant Velocity Model (CVM)}: A widely used linear kinematics model in vehicle tracking with an assumption of constant velocity. This model can also be generalized to forecasting trajectories for pedestrians.
    \item \textbf{Linear Regression (LR)}: A linear predictor which minimizes the least square error over the historical trajectories.
    \item \textbf{Probabilistic LSTM (P-LSTM)} \cite{Jiachen_ICRA19}: The model structure is the same as a vanilla LSTM. But a noise term sampled from the normal distribution is added in the input to incorporate uncertainty, which results in a probabilistic model.
    \item \textbf{Social-Forces} \cite{luber2010people}: The model is based on the concepts developed in the cognitive and social science communities that describe individual and collective pedestrian dynamics.
    \item \textbf{Social LSTM (S-LSTM)} \cite{alahi2016social}: The model encodes the trajectories with an LSTM layer whose hidden states serve as the input of a social pooling layer.
    \item \textbf{Social GAN (S-GAN and S-GAN-P)} \cite{gupta2018social}: The model introduces generative adversarial learning scheme into S-LSTM to improve performance.
    \item \textbf{Social Attention (S-ATT)} \cite{vemula2018social}: The model deals with spatio-temporal graphs with recurrent neural networks, which is based on the architecture of Structural-RNN \cite{jain2016structural}.
    \item \textbf{Clairvoyant attentive recurrent network (CAR-Net)} \cite{sadeghian2018car}: The model employs a physical attention module to capture agent-space interaction but without considering interactions among agents.
    \item \textbf{Social-Ways} \cite{amirian2019social}: The model uses a generative adversarial network (Info-GAN) to sample plausible predictions for any agent in the scene. 
    \item \textbf{SoPhie} \cite{sadeghian2019sophie}: The model leverages both context images and trajectory information to generate paths compliant to social and physical constraints.
    \item \textbf{DESIRE} \cite{lee2017desire}: The model is a deep stochastic inverse optimal control framework based on RNN encoders and decoders.
\end{itemize}

\end{document}